\documentclass[runningheads]{llncs}

\usepackage{eccv}

\usepackage{eccvabbrv}

\usepackage{graphicx}
\usepackage{booktabs}

\usepackage{tikz}
\usepackage{wrapfig}
\usepackage{amsmath}
\usepackage{amssymb}
\usepackage{geometry}
\usepackage{multirow}
\usepackage{makecell}
\usepackage{colortbl}
\usepackage{tikz-qtree}
\usepackage[misc]{ifsym}
\usepackage{fontawesome}
\usepackage{amsmath,amssymb}
\usepackage[export]{adjustbox}
\definecolor{MyRed}{HTML}{DC3977}
\definecolor{MyGreen}{HTML}{089099}
\usetikzlibrary{fit, calc, decorations.pathreplacing, positioning, shapes.geometric}
\newcommand\blfootnote[1]{
  \begingroup
  \renewcommand\thefootnote{}\footnote{#1}
  \addtocounter{footnote}{-1}
  \endgroup
}

\usepackage[accsupp]{axessibility}  

\usepackage{hyperref}

\usepackage{orcidlink}

\begin{document}

\title{PiTe: Pixel-Temporal Alignment for \\Large Video-Language Model} 


\author{
    Yang Liu$^{1,2}$\textsuperscript{$\star$} \and
    Pengxiang Ding$^{1}$\textsuperscript{$\star$} \and
    Siteng Huang$^{1}$ \and
    Min Zhang$^{1}$ \and \\
    Han Zhao$^{1}$ \and
    Donglin Wang$^{1}$\textsuperscript{\Letter}
}

\authorrunning{Y.~Liu, P. Ding et al.}

\institute{
    $^{1}$Westlake University \quad $^{2 }$Soochow University \\
    \email{\{liuyang67, dingpengxiang, wangdonglin\}@westlake.edu.cn}
    \blfootnote{\textsuperscript{$\star$} Equal contribution. \quad \textsuperscript{\Letter} Corresponding author.}
}

\maketitle


\begin{abstract}
Fueled by the Large Language Models (LLMs) wave, Large Visual-Language Models (LVLMs) have emerged as a pivotal advancement, bridging the gap between image and text.
However, video making it challenging for LVLMs to perform adequately due to the complexity of the relationship between language and spatial-temporal data structure.
Recent Large Video-Language Models (LVidLMs) align feature of static visual data like image into latent space of language feature, by general multi-modal tasks to leverage abilities of LLMs sufficiently.
In this paper, we explore fine-grained alignment approach via object trajectory for different modalities across both spatial and temporal dimensions simultaneously.
Thus, we propose a novel LVidLM by trajectory-guided \textbf{Pi}xel-\textbf{Te}mporal Alignment, dubbed \textbf{PiTe}, that exhibits promising applicable model property.
To achieve fine-grained video-language alignment, we curate a multi-modal pre-training dataset PiTe-143k, the dataset provision of moving trajectories in pixel level for all individual objects, that appear and mention in the video and caption both, by our automatic annotation pipeline.
Meanwhile, \textbf{PiTe} demonstrates astounding capabilities on myriad video-related multi-modal tasks through beat the state-of-the-art methods by a large margin.
\end{abstract}

\section{Introduction}

Large Language Models (LLMs) have rapidly gained popularity within the AI community, demonstrating astounding capabilities across a wide array of natural language tasks\cite{conf/nips/BrownMRSKDNSSAA20,conf/nips/Ouyang0JAWMZASR22,journals/corr/abs-2302-13971,journals/corr/abs-2307-09288,conf/acl/DuQLDQY022,vicuna2023}.
The powerful language comprehension abilities of LLMs drive researchers to explore their utility in addressing a broader spectrum of tasks across various domains.
Consequently, an increasing number of studies are focusing on developing comprehensive Large Visual-Language Models (LVLMs) to tackle vision-related tasks in zero-shot settings\cite{conf/icml/0008LSH23,zhu2024minigpt,zhang2022tree,conf/nips/Dai0LTZW0FH23}, particularly in the video understanding\cite{conf/nips/YangMSLS22,zhang2024llamaadapter,conf/ecai/LiuPZSLZ23,journals/corr/abs-2305-06355,conf/emnlp/ZhangLB23,maaz2024acl,journals/corr/abs-2311-13435}.
The pursuit of generalist Large Video-Language Models (LVidLMs) will be a perennial challenge.
Success in this endeavor hinges on effectively leveraging the exceptional understanding, reasoning, and generative capacities inherently present in LLMs.

\begin{figure}[!t]
    \centering
    \subfloat[Illustration of training paradigm comparison.]{
        \label{fig:paradigm}
        \includegraphics[width=0.6\linewidth,trim=15 5 5 5,clip]{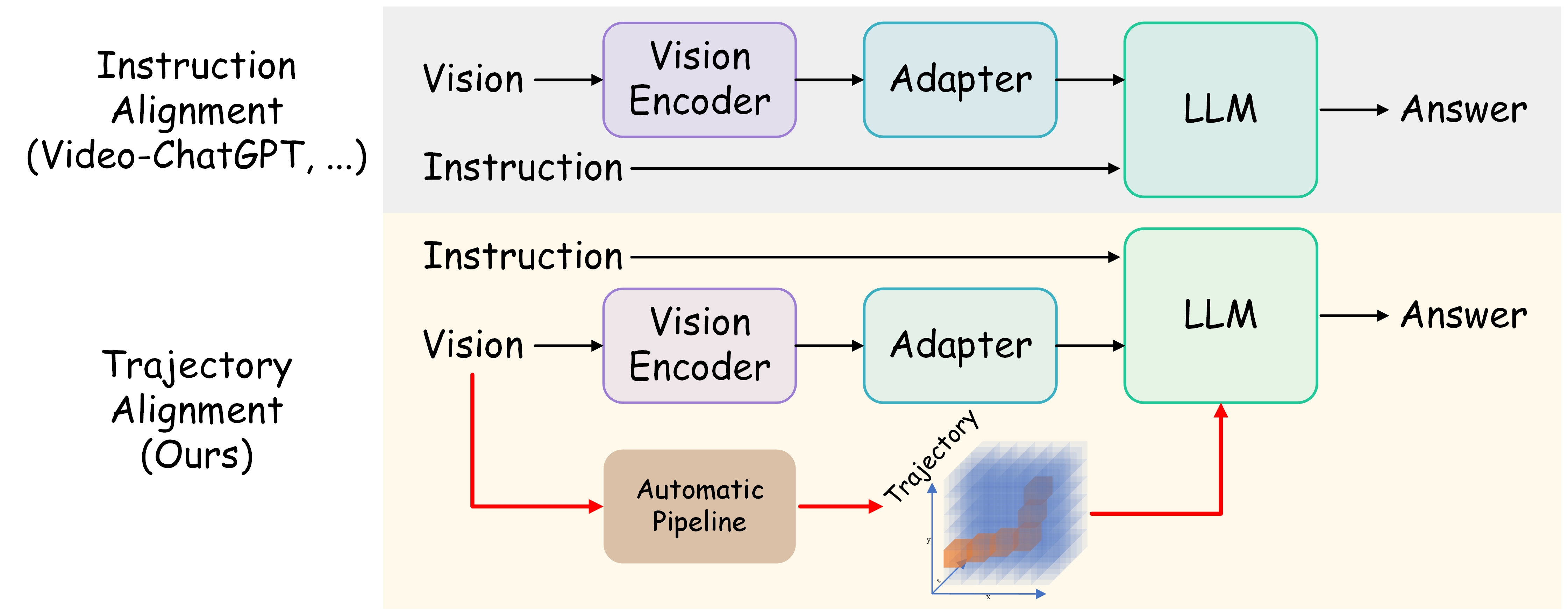}
    }
    \hfill
    \subfloat[Performance comparison. ]{
        \label{fig:radar}
        \includegraphics[width=0.33\linewidth,trim=0 0 0 15,clip]{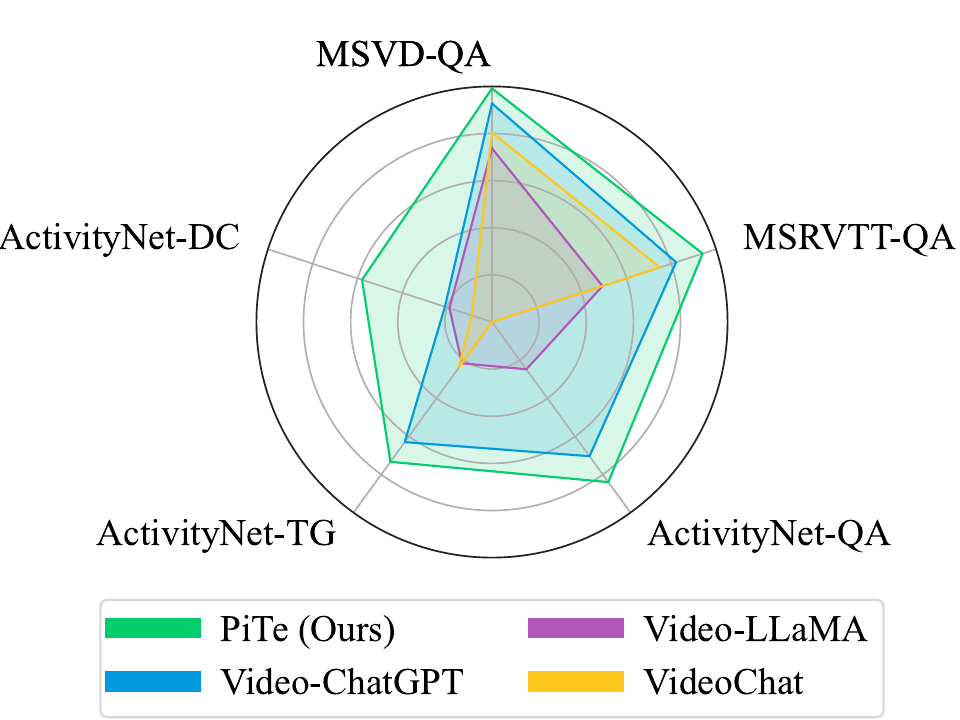}
    }
    \caption{Comparison with existing LVidLMs in terms of alignment paradigm and performance. For Fig.~\ref{fig:radar}, \textit{QA}, \textit{TG}, \textit{DC} denote question answering, temporal grounding and dense captioning, respectively.}
    \label{fig:intro}
    \vspace{-2ex}
\end{figure}


One potential route towards addressing the issue is aligning visual feature into latent space of language feature.
To achieve this, existing LVidLMs apply large-scale vanilla instruction tuning\cite{zhang2024llamaadapter,maaz2024acl,journals/corr/abs-2305-06355,conf/emnlp/ZhangLB23}.
However, the conventional question-answering training paradigm primarily assists LLMs in understanding visual data from a spatial perspective, posing challenges in effectively capturing temporal dynamics and spatial consistency relationships.
Therefore, relying solely on instruction tuning proves insufficient for achieving comprehensive video comprehension, given the intricate spatial-temporal data structure involved.
It is crucial to align different modalities across both spatial and temporal dimensions.
Furthermore, offering more fine-grained cross-modal alignment guidance significantly enhances LVidLMs' ability to comprehend videos\cite{conf/icmr/Liu24CVLA}.

To bridge the gap, we introduce a novel LVidLM named \textbf{PiTe}, which emploies trajectories to intricately align vision and language across both spatial and temporal dimensions at the pixel level, and the distinction from conventional approaches is illustrated in Figure~\ref{fig:paradigm}.
By requiring the model to forecast the trajectory of individual objects mentioned in the text within the video, it enables the learning of fine-grained text-to-pixel alignment through exploiting the video context along the temporal dimension and enhancing its ability to generate output based on evidence.

Subsequently, due to there are no ready-made video-language dataset with moving trajectory of objects, we curate a large-scale video-language dataset PiTe-143k through an automated annotation pipeline.
Consequently, as shown in Figure~\ref{fig:radar}, the proposed \textbf{PiTe} significantly augments the LVidLM's capacity to understand videos comprehensively, leading to promising, competitive, and state-of-the-art performance in question-answering, temporal grounding, and dense captioning tasks under zero-shot conditions.

Overall, our principal contributions in this paper are summarized as follows: \\
$\bullet$ We curate a large scale video-language dataset PiTe-143k with trajectory for all individual objects by automatic annotation pipeline. \\
$\bullet$ We propose a novel LVidLM \textbf{PiTe} that utilize trajectory to align video and language features across both spatial and temporal dimensions. \\
$\bullet$ Extensive experimental results and analysis on myriad datasets for zero-shot video question answering, temporal grounding, and dense captioning tasks demonstrate the superiority of \textbf{PiTe}.
\section{Related Work}

\subsection{Large Language Models}

Over the last few years, pioneering foundation language models like GPT-1\cite{Radford2018ImprovingLU}, BERT\cite{conf/naacl/DevlinCLT19}, GPT-2\cite{Radford2019LanguageMA}, and T5\cite{journals/jmlr/RaffelSRLNMZLL20} laid the groundwork, but GPT-3\cite{conf/nips/BrownMRSKDNSSAA20} groundbreaking model parameters to 175 billion size to achieve remarkable zero-shot performance.
Besides, research on scaling law\cite{journals/corr/abs-2001-08361} has steered language models to a larger scale.
Therefore, driven by the success of InstructGPT\cite{conf/nips/Ouyang0JAWMZASR22} and ChatGPT\cite{OpenAI_ChatGPT} which training by reinforcement learning with human feedback (RLHF) based on GPT-3, Large Language Models (LLMs) has made waves in the natural language processing (NLP) community due to its capabilities in language understanding, logical reasoning, and generation.
The GPT's success suggest a promising path towards building LLMs, several open-source LLMs have been proposed following it with similar performance including OPT\cite{journals/corr/abs-2205-01068}, BLOOM\cite{journals/corr/abs-2211-05100}, GLM\cite{conf/acl/DuQLDQY022}, LLaMA\cite{journals/corr/abs-2302-13971,journals/corr/abs-2307-09288}, and Vicuna\cite{vicuna2023}.
Our investigation delves into leveraging the striking language comprehension and zero-shot generalization abilities of LLMs beyond the confines of linguistic modalities.
Specifically, we aim to extend these capabilities to multi-modal scenarios, thereby exploring their potential in processing diverse forms of information across different modalities.

\subsection{Large Visual-Language Models}

The surge of LLMs has lead to major advancements in NLP tasks, and also has incited interest in developing Large Visual-Language Models (LVLMs).
Building a unify LLM with visual inputs for visual language tasks thus remains one of the most important desiderata for LVLMs.
Flamingo\cite{conf/nips/AlayracDLMBHLMM22} and OpenFlamingo\cite{journals/corr/abs-2308-01390} fuse visual information into intermediate embedding for a frozen LLM by cross-attention mechanism, and train on billions of image-text pairs to align visual and linguistic modalities.
Similarly, BLIP-2\cite{conf/icml/0008LSH23} introduced the concept of Q-Former to align visual features more effectively with language space.
Moreover, MiniGPT-4\cite{zhu2024minigpt} enhances its usability significantly by further fine-tuning on more detailed image descriptions with just one projection layer to align a frozen visual encoder with a frozen LLM, and the LLaVA series\cite{liu2023visual,Liu_2024_CVPR} use simply a multi-layer perception  (MLP) in place of the Q-Former and two-stage instruction tuning to enhance this process.
Furthermore, PixelLLM\cite{Xu_2024_CVPR} leverage the location coordinate of every word in the caption in the image as the connection between different modalities to strengthen the model's performance for the object detection task.
Our primary focus lies in transferring the exceptional language comprehension capabilities of LLMs to the analysis of dynamic, continuous visual data found in videos, as opposed to static visual data such as images.

\subsection{Large Video-Language Models}

Recently, many efforts have been made to transfer the task-handling capability of LVLMs to the video modality, leading to the emergence of Large Video-Language Models (LVidLMs) like VideoChat\cite{journals/corr/abs-2305-06355}, Video-LLaMA\cite{conf/emnlp/ZhangLB23}, and Video-ChatGPT\cite{maaz2024acl}.
Prior researches have demonstrated the capability of LLMs to perform diverse tasks on video content, guided by user instructions through a two-stage training process.
These studies align static visual features with LLMs, followed by instruction tuning on datasets annotated either by GPT or humans.
Despite being effective in video understanding, the lack of fine-grained spatial-temporal modeling in these models prevents them from understanding or locating object in detail or specific segments.
We propose a novel fine-grained alignment strategy at the pixel level across spatial and temporal dimensions to enhance the ability of LLMs to comprehensively analyze video content, thereby facilitating a more detailed understanding of the visual information presented.
\section{PiTe-143k Dataset}
\label{sec:dataset}

\begin{table}[!t]
    \caption{Comparison between PiTe-143k and existing video instruction datasets.}
    \label{tab:dataset}
    \centering
    \setlength{\tabcolsep}{4pt}
    \resizebox{\linewidth}{!}{
        \begin{tabular}{lcccccc}
            \toprule
            \multicolumn{1}{c}{\textbf{Dataset}} & \textbf{Total Dur.} & \textbf{Avg. Dur.} & \textbf{\#Videos} & \textbf{\#Events} & \makecell[c]{\textbf{Temporal}\\\textbf{Localization}} & \makecell[c]{\textbf{\#Objects}\\\textbf{Trajectories}} \\
            \midrule
            VideoChat\cite{journals/corr/abs-2305-06355} & 41h & 18s & 8.2k & {\color{MyRed}\faTimes} & {\color{MyRed}\faTimes} & {\color{MyRed}\faTimes} \\
            Valley\cite{journals/corr/abs-2306-07207} & 608h & 40s & 54.7k & {\color{MyRed}\faTimes} & {\color{MyRed}\faTimes} & {\color{MyRed}\faTimes} \\
            Video-ChatGPT\cite{maaz2024acl} & 432h & \textbf{117s} & 13.3k & {\color{MyRed}\faTimes} & {\color{MyRed}\faTimes} & {\color{MyRed}\faTimes} \\
            \midrule
            \textbf{PiTe-143k (Ours)} & \textbf{2086.44h} & 52.18s & \textbf{143.64k} & \textbf{343.93k} & {\color{MyGreen}\faCheck} & 1.02M \\
            \bottomrule
        \end{tabular}
    }
\end{table}

To facilitate fine-grained multi-modal alignment research at the pixel level, we introduce a large-scale video-language dataset PiTe-143k.
This dataset fills a crucial gap in the existing resources by providing extensive object moving trajectories with video instruction, which were previously unavailable in ready-made datasets.
PiTe-143k constructed based on InternVid-10M-FLT\cite{wang2024internvid,Huang_2024_CVPR} that each instance contains and entire video and multiple clip captions with start-stop timestamps.
As shown in Table~\ref{tab:dataset}, PiTe-143k comprises 343.93 thousand event segments and 1.02 million moving trajectory for all individual objects that appear in both visual and textual modalities.
To facilitate this objective, we establish an automatic annotation pipeline for PiTe-143k, fostering the advancement of LVidLMs for nuanced pixel-level video comprehension.

\begin{figure}[!t]
    \centering
    \includegraphics[width=\linewidth]{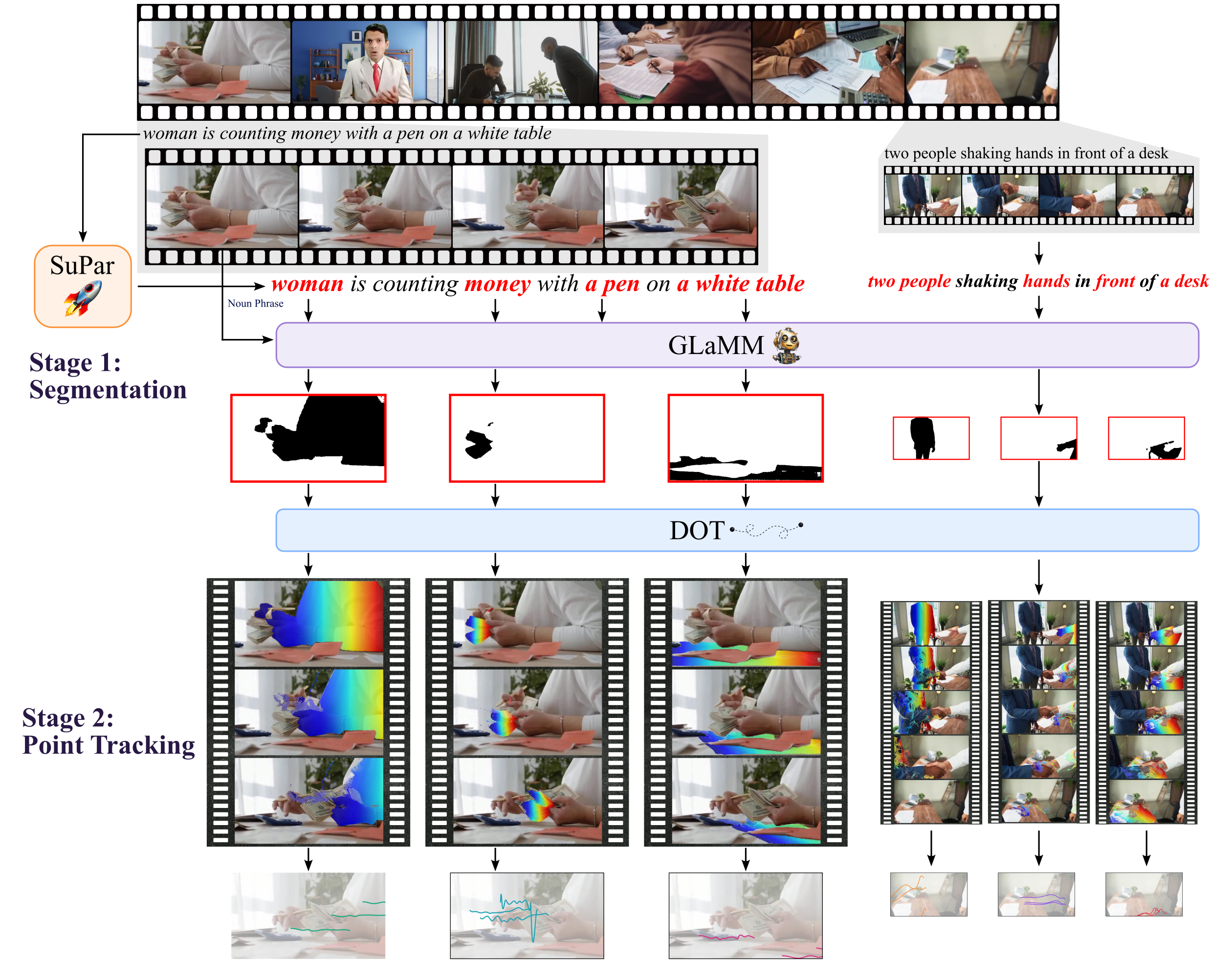}
    \caption{Automatic annotation pipeline for PiTe-143k. The video sample in the figure showcases two events positioned at the commencement and conclusion of the video. The procedure for extracting noun phrases by SuPar\cite{conf/acl/ZhangLZ20,conf/ijcai/ZhangZL20} is elucidated in Fig.~\ref{fig:tree}.}
    \label{fig:pite_data}
\end{figure}

The automatic annotation pipeline for PiTe-143k comprises two primary stages, as depicted in Fig.~\ref{fig:pite_data}: (1) Stage 1 involves the noun phrases extraction and referring expressions segmentation, thereby generating object masks within the frame for all individual objects referenced in the event caption; (2) Stage 2 centers on point tracking to capture the moving trajectories corresponding to the masks obtained in Stage 1.

\subsection{Referring Expression Segmentation}

In stage 1, we aim to build closely fine-grained connection between video and language.
To this end, we extract all noun phrases from caption and find the corresponding objects in the clip.

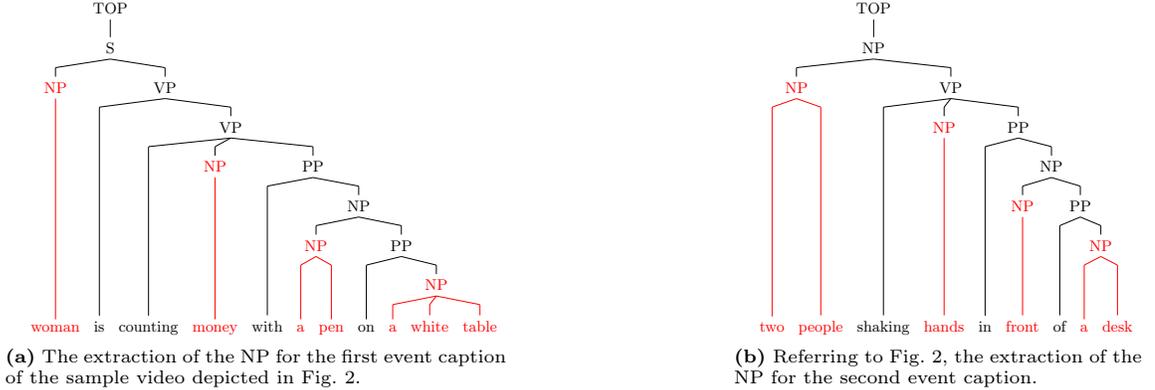
\begin{figure}[!t]
    \centering
    \subfloat[The extraction of the NP for the first event caption of the sample video depicted in Fig.~\ref{fig:pite_data}.]{
        \label{fig:money-tree}
        \begin{tikzpicture}[
            , scale=0.65
            , level distance=23pt
            , frontier/.style={distance from root=185pt}
            , every tree node/.style={align=center,anchor=base}
            , edge from parent/.style={draw,edge from parent path={(\tikzparentnode.south) {[rounded corners=0.5pt]-- ($(\tikzchildnode |- \tikzparentnode.south) + (0, -5pt)$) -- (\tikzchildnode)}}}
        ]
        \Tree
        [.TOP
            [.S
                [.\node[text=red]{NP}; \edge[draw=red];\node[text=red]{woman}; ]
                [.VP
                    is
                    [.VP
                        counting
                        [.\node[text=red]{NP}; \edge[draw=red];\node[text=red]{money}; ]
                        [.PP
                            with
                            [.NP
                                [.\node[text=red]{NP};
                                    \edge[draw=red];\node[text=red]{a};
                                    \edge[draw=red];\node[text=red]{pen}; ]
                                [.PP
                                    on
                                    [.\node[text=red]{NP};
                                        \edge[draw=red];\node[text=red]{a};
                                        \edge[draw=red];\node[text=red]{white};
                                        \edge[draw=red];\node[text=red]{table}; ]
                                ]
                            ]
                        ]
                    ]
                ]
            ]
        ];
        \end{tikzpicture}
    }
    \hfill
    \subfloat[Referring to Fig.~\ref{fig:pite_data}, the extraction of the NP for the second event caption.]{
        \label{fig:hand-tree}
        \begin{tikzpicture}[
            , scale=0.65
            , level distance=23pt
            , frontier/.style={distance from root=185pt}
            , every tree node/.style={align=center,anchor=base}
            , edge from parent/.style={draw,edge from parent path={(\tikzparentnode.south) {[rounded corners=0.5pt]-- ($(\tikzchildnode |- \tikzparentnode.south) + (0, -5pt)$) -- (\tikzchildnode)}}}
        ]
        \Tree
        [.TOP
            [.NP
                [.\node[text=red]{NP};
                    \edge[draw=red];\node[text=red]{two};
                    \edge[draw=red];\node[text=red]{people}; ]
                [.VP
                    shaking
                    [.\node[text=red]{NP}; \edge[draw=red];\node[text=red]{hands}; ]
                    [.PP
                        in
                        [.NP
                            [.\node[text=red]{NP}; \edge[draw=red];\node[text=red]{front}; ]
                            [.PP
                                of
                                [.\node[text=red]{NP};
                                    \edge[draw=red];\node[text=red]{a};
                                    \edge[draw=red];\node[text=red]{desk}; ]
                            ]
                        ]
                    ]
                ]
            ]
        ];
        \end{tikzpicture}
    }
    \caption{Two samples of constituency parser for Noun Phrase (NP) extraction.}
    \label{fig:tree}
    \vspace{-2ex}
\end{figure}

At inception, we leverage constituency parser SuPar\cite{conf/acl/ZhangLZ20,conf/ijcai/ZhangZL20} for language to extract noun phrase as shown in Fig.~\ref{fig:tree}.
Notably, in order to pass the simplest and most straightforward language instructions in next step, we only extract noun phrase from the lowest layer.
For example in Fig.~\ref{fig:money-tree}, we consider two noun phrases \textit{a pen} and \textit{a white table}, but the parent node of the former that denotes \textit{a pen on a white table} not in our consideration because of the complexity of its composition.
Following this, we utilize GLaMM\cite{Rasheed_2024_CVPR}, the first LVLM that can generate natural language responses seamlessly intertwined with corresponding object segmentation masks, to obtain the corresponding segmentation mask in the first frame of the clip for the text-based referring expression.
While certain objects in the video, such as \textit{a pen} as illustrated in Fig.\ref{fig:money-tree} and Fig.\ref{fig:pite_data}, may be too small to accurate detection.
In such challenging cases, we disregard the trajectory information of the noun phrase.
Despite this limitation, it has a minimal impact on the overall performance when utilizing extensive pre-training data on a large scale.
Meanwhile, leveraging the exceptional language comprehension capabilities of LLMs, GLaMM can effectively filter out invalid referring expressions, those that do not constitute legal object references, such as \textit{front} as depicted in Fig.~\ref{fig:hand-tree}.

\subsection{Point Tracking}

In stage 2, we aim to transfer the connection constructed in the previous stage to video, expanding out the temporal dimension specific to video compared to image.
To this end, we track all individual objects in their clip to obtain the trajectory, the trajectory indicates the connection between video and language in both spatial and temporal dimensions.

The stage 2 commences when we employ DOT\cite{Le_Moing_2024_CVPR}, a simple-yet-efficient method for tracking point to recover the trajectory of any scene point, for each clip to capture the trajectories for any point in first frame.
According to our observation, the caption of each clip mainly describes simple video content in short sentences, so most of the caption corresponds to just one scene clip, which enables us to track objects that identified in the first frame.
Subsequently, filter trajectories according to the segmentation mask of objects obtained in stage 1.
So far, we obtain the trajectories for all objects in each clip for each video, we create the connection between video and language from both spatial and temporal through the trajectories, the existence of trajectories in video denotes whether the object exist, and the value of trajectory represent where the object exist in video.
Lastly, we utilize the k-means$++$\cite{conf/soda/ArthurV07} clustering algorithm to condense trajectories into three key points, effectively reducing computational demands.
This approach is founded on the premise that three points adequately capture the typical geometric shape of objects, striking a balance between precision and computational efficiency.
Furthermore, we conduct a comparative analysis of the performance using various key tracking points, as discussed in Section~\ref{sec:n_point}.
\section{PiTe}

\begin{figure}[!t]
    \centering
    \includegraphics[width=\linewidth]{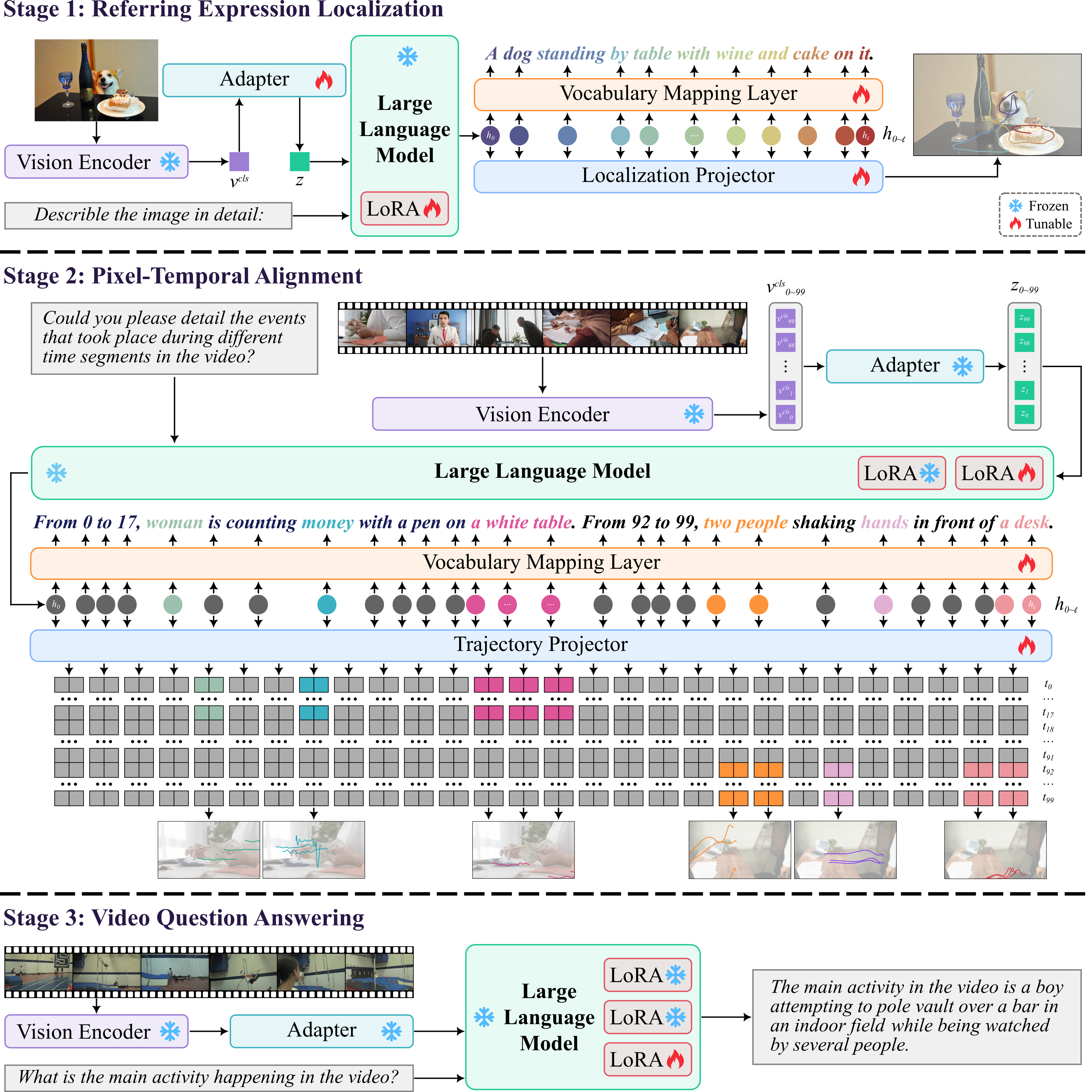}
    \caption{Schematic of \textbf{PiTe} framework for video-language alignment.}
    \label{fig:framework}
\end{figure}

In this section, we propose a novel Large Video-Language Model (LVidLM), \textbf{PiTe}, which align video and language by trajectories across both spatial and temporal dimension.
Fig.~\ref{fig:framework} illustrates an overview of \textbf{PiTe}.

\subsection{Architecture}

\textbf{PiTe} is composed of a vision encoder to encode frames from video implemented as ViT\cite{conf/iclr/DosovitskiyB0WZ21}, a vision adapter to project visual feature to semantic space of LLMs implemented as a linear projection layer, a LLM Vicuna v1.5\cite{vicuna2023}, and a localization projector or trajectory projector in separate training stage to guide LLMs to understand visual information implemented as a linear projection.

\subsubsection{Vision Encoder.}

Raw video data can be expressed as multiple frames such as $\textbf{v}=\{f_{1}, f_{2}, \dots, f_{N}\}\in\mathbb{R}^{N\times H\times W\times C}$ (frames$\times$ height $\times$ width $\times$ channels).
Following previous studies\cite{liu2023visual,conf/emnlp/ZhangLB23,maaz2024acl}, we adopt ViT-L/14\cite{conf/iclr/DosovitskiyB0WZ21} pre-trained from CLIP\cite{conf/icml/RadfordKHRGASAM21} as the vision encoder $\textrm{ViT}$ to encode visual data.
We uniformly sample $N$ frames for video $\textbf{v}$, and encode $i$-th frame $f_{i}$ through the vision encoder $\textrm{ViT}$:
\begin{small}
\begin{equation}
\left\{v_{i}^{cls}, v_{i}^{1}, v_{i}^{2}, \dots, v_{i}^{P}\right\}=\textrm{ViT}\left(f_{i}\right),
\label{eq:vit}
\end{equation}
\end{small}
where $P$ denotes the number of patches in the vision encoder $\textrm{ViT}$.

\subsubsection{Visual Adapter.}

A simpler projector forces the LLMs to learn more on handling visual inputs, leading to better generalization\cite{Lin_2024_CVPR}.
Hence, we utilize the global feature $v_{i}^{cls}$ from vision encoder $\textrm{ViT}$ as the representation for the $i$-th frame $f_{i}$, and we apply a linear projection layer $\mu(\cdot)$ to connect the frame feature into the word embedding space of LLMs:
\begin{small}
\begin{equation}
z_{i}=\mu\left(v_{i}^{cls}\right).
\label{eq:adapter}
\end{equation}
\end{small}
Subsequently, a sequence of frame tokens $\textbf{z}=\{z_{1}, z_{2}, \dots, z_{N}\}\in\mathbb{R}^{N\times d}$ becomes the input that LLMs can understand, $d$ denotes the hidden dimension of LLMs.

\subsubsection{Large Language Model.}

After we tokenize and encode video into frame tokens $\textbf{z}$, we concatenate it with textual tokens $\textbf{w}=\{w_{1}, w_{2}, \dots, w_{L}\}\in\mathbb{R}^{L}$ and feed as the input to LLMs, we treat visual input as a foreign language in this process.
Based on this, LLMs can further encode the input sequence to understand the video and text, then reasoning and generation the response using auto-regressive decoding as follows:
\begin{gather}
h_{i}=\textrm{LLM}^{-}\left(\textbf{z},\textbf{w}_{1:i-1}\right), \\
w_{i}=\textrm{argmax}\left(\textbf{m}_{v}\cdot h_{i}\right),
\end{gather}
where $\textrm{LLM}^{-}$ denotes the LLM without the last vocabulary mapping layer, $h_{i}$ denotes the hidden states of $i$-th token generated by $\textrm{LLM}^{-}$, $\textbf{m}_{v}\in\mathbb{R}^{\mid V\mid\times d}$ denotes the weight of the linear vocabulary mapping layer.

\subsection{Training Strategy}

For \textbf{PiTe} model training, we consider a three stage instruction-tuning procedure, as depicted in Fig.~\ref{fig:framework}: (1) Stage 1 centered around training adapters using image-caption pairs; (2) Stage 2 is focuses on aligning video and language features through trajectories; (3) Stage 3 is dedicated to enhancing the model's comprehension by following human instructions through high-quality dialogue instruction tuning.

\subsubsection{Stage 1: Referring Expression Localization.}

At the initial stage, we aim to train the visual adapter that can align visual features with semantic space of LLMs.
To this end, we employ Localized Narratives dataset\cite{conf/eccv/Pont-TusetUCSF20} that contains annotations of human annotators narrating a given image, together with a mouse trajectory of the annotators’ attention during the narration.
This gives synchronized locations for all words in the narration sentence, the cross-modal attention of human can be used to train our model as condition to bridge vision and language.

There is only one visual tokens $\textbf{z}=\{z_{1}\}\in\mathbb{R}^{1\times d}$ in this training stage for image instead of video, this can align vision with language in spatial to train adapter without consider temporal information.
To use the same language features for localization, we simply add a multi-layer perception (MLP) as localization projector $\varphi(\cdot)$ parallel with the vocabulary mapping layer, which maps the language feature to a 2-dimension location:
\begin{small}
\begin{equation}
p_{i}=\varphi(h_{i}),
\end{equation}
\end{small}
where $p_{i}$ denotes the predicted coordinate for textual token $w_{i}$.

Overall, the leaning objective in the first stage was calculated by standard label-smoothed cross-entropy loss to train captioning output, and $L_{1}$ regression loss to train the localization output:
\begin{small}
\begin{equation}
\mathcal{L}_{1}=\frac{1}{\ell}\sum_{i=1}^{\ell}\left(\textrm{CE}\left(\textrm{LLM}\left(\textbf{z}, \textbf{w}_{1:i-1}\right), w_{i}\right)+\lambda\mid\hat{p}_{i}-p_{i}\mid\right),
\end{equation}
\end{small}
where $\ell$ is the length of the generated sequence, $\hat{p}_{i}$ represents the ground truth of location, and $\textrm{CE}(\cdot,\cdot)$ denotes the cross-entropy function.
To enhance training efficiency, we utilize LoRA\cite{conf/iclr/HuSWALWWC22} for fine-tuning the LLM.

\subsubsection{Stage 2: Pixel-Temporal Alignment.}

After stage 1, the LLM model becomes proficient in understanding visual information.
In the stage 2, we aim to train the LLM to understand sequential frames in video.
To achieve this target, we curate a detailed object tracking dataset PiTe-143k, as described in Section~\ref{sec:dataset} that uses trajectory as condition to bridge vision and language across both spatial and temporal dimensions.
Therefore, the alignment guidance in pixel level improves the model's video fine-grained understanding reliability and overall usability.

Similar to stage 1, we use the same language features for alignment by a MLP as trajectory projector $\rho(\cdot)$ to map the language feature to a 2-dimension location:
\begin{small}
\begin{equation}
\textbf{p}_{i}=\rho(h_{i}),
\end{equation}
\end{small}
where $\textbf{p}_{i}$ denotes the trajectory matrix in $P$ points and $N$ frames for textual token $w_{i}$.
Here, we define $p_{ijk}$ indicates the coordinate for token $w_{i}$ for $i$-th point for model tracking in frame $f_{j}$.

Overall, the leaning objective in stage 2 was calculated by standard label-smoothed cross-entropy loss to train the generation output, and $L_{1}$ regression loss to train the trajectory output as the condition:
\begin{small}
\begin{equation}
\mathcal{L}_{2}=\frac{1}{\ell}\sum_{i=1}^{\ell}\left(\textrm{CE}\left(\textrm{LLM}\left(\textbf{z}, \textbf{w}_{1:i-1}\right), w_{i}\right)+\frac{\lambda}{P\cdot N}\sum_{j=1}^{P}\sum_{k=1}^{N}\mid\hat{p}_{ijk}-p_{ijk}\mid\right),
\end{equation}
\end{small}
where $P$ is the number of the points for model tracking to generate trajectory, and $\textbf{z}\in\mathbb{R}^{N\times d}$ represents the sequence of visual embedding.
We merge the LoRA trained in the stage 1 with the original model and introduce a new LoRA module.

It is notable that we use localization projector $\varphi(\cdot)$ trained in previous stage to initialize the trajectory projector $\rho(\cdot)$.
Specifically, we define the weight of localization projector $\varphi(\cdot)$ and trajectory projector $\rho(\cdot)$ as $\textbf{m}_{\varphi}\in\mathbb{R}^{P\cdot N\cdot 2\times d}$ and $\textbf{m}_{\rho}\in\mathbb{R}^{2\times d}$, respectively.
localization projector $\varphi(\cdot)$ maps a 2-dimension coordinate on the input image for each token of the LLM output, as for trajectory projector $\rho(\cdot)$, it also output 2-dimension coordinates, but more than $P\cdot N$ times for $P$ points to tracking for $N$ frames.
For each point of each frame, the parameter of trajectory projector $\rho(\cdot)$ initialized by the localization projector $\varphi(\cdot)$:
\begin{small}
\begin{equation}
\textbf{m}_{\varphi}=\overbrace{\textbf{m}_{\rho}\oplus\textbf{m}_{\rho}\oplus\cdots\oplus\textbf{m}_{\rho}}^{P\cdot N},
\end{equation}
\end{small}
where $\oplus$ denotes concatenation of matrix in first dimension.

\begin{table}[!t]
    \caption{Training hyper-parameters of PiTe.}
    \label{tab:hyperparam}
    \centering
    \resizebox{\linewidth}{!}{
        \begin{tabular}{lccclccc}
            \toprule
            \multicolumn{1}{c}{\textbf{Configuration}} & \textbf{Stage 1} & \textbf{Stage 2} & \textbf{Stage 3} & \multicolumn{1}{c}{\textbf{Configuration}} & \textbf{Stage 1} & \textbf{Stage 2} & \textbf{Stage 3} \\
            \midrule
            Vision Encoder & \multicolumn{3}{c}{OpenAI-CLIP-L/14} & Learning Rate & \multicolumn{3}{c}{0.0001} \\
            Image/Frame Resolution & \multicolumn{3}{c}{224$\times$224} & LoRA & \multicolumn{3}{c}{r=64 \& $\alpha$=128} \\
            Adapter Parameter & Tunable & Frozen & Frozen & Numerical Precision & \multicolumn{3}{c}{BFloat16} \\
            Video Frames & \multicolumn{3}{c}{100} & Epoch & 1 & 2 & 2 \\
            LLM & \multicolumn{3}{c}{Vicuna-7B/13B-v1.5} & Global Batch Size & \multicolumn{3}{c}{256} \\
            LLM Sequence Length & \multicolumn{3}{c}{2048} & Learning Rate Schedule & \multicolumn{3}{c}{Cosine Decay} \\
            Optimizer & \multicolumn{3}{c}{AdamW} & Warm-up Ratio & \multicolumn{3}{c}{0.03} \\
            \bottomrule
        \end{tabular}
    }
\end{table}

Beyond trajectories, our model is attuned to temporal boundaries within the generated text.
Specifically, we structure the generation as $\texttt{\dots, from s to e}$ or $\texttt{From s to e, \dots}$ to facilitate the model to learn in temporal dimension.
Here, $\texttt{\dots}$ encapsulates the event description, while $\texttt{s}$ and $\texttt{e}$ denote the frame indexes corresponding to the start and end timestamps of event, respectively.
This approach further augments the model's understanding of temporal boundaries\cite{Huang_2024_CVPR}.

Dissimilar to the initial training stage, not all generated words are associated with trajectories.
In cases where objects lack trajectories or vanish from view over time, we uniformly assign the coordinates of their ground truth as $(-1, -1)$ to signify their absence.

\subsubsection{Stage 3: Video Question Answering.}

Following stage 2, we incorporate high-quality dialogue data Valley\cite{journals/corr/abs-2306-07207} and Video-ChatGPT\cite{maaz2024acl} in one turn for instruction tuning, enabling the model to follow human instructions for more accurate and generalize capabilities of video understanding.

The leaning objective in third stage was calculated by standard label-smoothed cross-entropy loss for auto-regression generation:
\begin{small}
\begin{equation}
\mathcal{L}_{3}=\frac{1}{\ell}\sum_{i=1}^{\ell}\textrm{CE}\left(\textrm{LLM}\left(\textbf{z}, \textbf{w}_{1:i-1}\right), w_{i}\right).
\end{equation}
\end{small}
Similar to stage 2, we merge the LoRA trained in the stage 1 and stage 2 with the original model and introduce a new LoRA module.
\section{Experiments}

\begin{table}[!t]
    \caption{Comparison between different LVidLMs on zero-shot question-answer.}
    \label{tab:main_zsvqa}
    \centering
    \resizebox{\linewidth}{!}{
        \begin{tabular}{lccccccc}
            \toprule
            \multicolumn{1}{c}{\multirow{2}{*}{\textbf{Model}}} & \multirow{2}{*}{\textbf{LLM Size}} & \multicolumn{2}{c}{\textbf{MSVD-QA}\cite{conf/mm/XuZX0Z0Z17}} & \multicolumn{2}{c}{\textbf{MSRVTT-QA}\cite{conf/cvpr/XuMYR16}} & \multicolumn{2}{c}{\textbf{ActivityNet-QA}\cite{conf/aaai/YuXYYZZT19}} \\
            \cmidrule(lr){3-4} \cmidrule(lr){5-6} \cmidrule(lr){7-8}
             &  & Accuracy$\uparrow$ & Score$\uparrow$ & Accuracy$\uparrow$ & Score$\uparrow$ & Accuracy$\uparrow$ & Score$\uparrow$ \\
            \midrule
            FrozenBiLM\cite{conf/nips/YangMSLS22} & 1B & 32.2 & - & 16.8 & - & 24.7 & - \\
            LLaMA-Adapter\cite{zhang2024llamaadapter} & 7B & 54.9 & 3.1 & 43.8 & 2.7 & 34.2 & 2.7 \\
            VideoChat\cite{journals/corr/abs-2305-06355} & 7B & 56.3 & 2.8 & 45.0 & 2.5 & - & 2.2 \\
            Video-LLaMA\cite{conf/emnlp/ZhangLB23} & 7B & 51.6 & 2.5 & 29.6 & 1.8 & 12.4 & 1.1 \\
            Video-ChatGPT\cite{maaz2024acl} & 7B & 64.9 & 3.3 & 49.3 & 2.8 & 35.2 & 2.7 \\
            PG-Video-LLaVA\cite{journals/corr/abs-2311-13435} & 7B & 64.1 & 3.7 & 51.6 & 3.3 & 39.9 & \textbf{3.3} \\
            \rowcolor[rgb]{ .949,  .949,  .949} \textbf{PiTe (Ours)} & 7B & \textbf{68.4} & \textbf{3.9} & \textbf{56.4} & \textbf{3.5} & \textbf{42.0} & \textbf{3.3} \\
            \midrule
            \rowcolor[rgb]{ .949,  .949,  .949} \textbf{PiTe (Ours)} & 13B & \textbf{71.6} & \textbf{4.0} & \textbf{57.7} & \textbf{3.5} & \textbf{42.2} & \textbf{3.4} \\
            \bottomrule
        \end{tabular}
    }
\end{table}

\subsection{Experimental Setup}

\subsubsection{Tasks, Datasets, and Evaluation Metrics.}

We conduct a quantitative evaluation of LVidLMs' video understanding capabilities across three tasks:
(1) Video Question Answering: This task assesses the comprehensive video comprehension abilities of LVidLMs by requiring the model to answer a variety of questions about the video content based on its understanding.
We perform this task on three datasets: MSVD-QA\cite{conf/mm/XuZX0Z0Z17}, MSRVTT-QA\cite{conf/cvpr/XuMYR16}, and ActivityNet-QA\cite{conf/aaai/YuXYYZZT19}.
The evaluation pipeline for video understanding follows Video-ChatGPT\cite{maaz2024acl}, and we report the accuracy and score, which is assessed using GPT-Assistant\cite{OpenAI_ChatGPT}.
(2) Video Temporal Grounding: This task evaluates LVidLMs' capacity to discern the starting and ending timestamps of a segment corresponding to the description of a video clip.
This task demands the model to effectively grasp the temporal aspects of the video.
We conduct this task on the ActivityNet Captions dataset\cite{conf/iccv/KrishnaHRFN17} and calculate Intersection over Union (IoU) between the model-generate time segments and the ground truth time segments.
We report mean IoU (mIoU) and Recall@1, IoU$\ge m$ (R@m) metric, where $m$ values are set at $\{0.3, 0.5, 0.7\}$.
(3) Video Dense Captioning: This task requires the model to produce all events depict in the video along with their corresponding start and end timestamps.
It necessitates the model to comprehend both the spatial and temporal dimensions of the video simultaneously.
We conduct this task on the ActivityNet Captions dataset\cite{conf/iccv/KrishnaHRFN17}.
Initially, we reported SODA\underline{~}c\cite{conf/eccv/FujitaHKON20}, followed by averages of CIDEr\cite{conf/cvpr/VedantamZP15} and METEOR\cite{conf/wmt/LavieA07} under different IoU thresholds of {0.3, 0.5, 0.7, 0.9} based on generate events and ground truth matched pairs to provide a comprehensive analysis.
In this paper, all experiments were conducted in a zero-shot setting, and higher values of all evaluation metrics indicate superior performance.

\subsubsection{Implementation Details.}

In this paper, we employ Vicuna v1.5\cite{vicuna2023} as LLM to train the \textbf{PiTe} model at two scales: 7B and 13B.
Leveraging the efficiency of LoRA\cite{conf/iclr/HuSWALWWC22}, the training of the 7B model can be completed in approximately 10 hours using a single Nvidia 8-A100 (80GB VRAM) node, while the 13B model requires around 17 hours.
More hyper-parameter settings are shwon in Table~\ref{tab:hyperparam}.
\footnote{Our dataset and code release at \href{https://github.com/yliu-cs/PiTe}{https://github.com/yliu-cs/PiTe}.}


\begin{table}[!t]
    \caption{Comparison between different LVidLMs in temporal video grounding and dense video captioning tasks on ActivityNet\cite{conf/cvpr/HeilbronEGN15}.}
    \label{tab:main_an}
    \centering
    \resizebox{\linewidth}{!}{
        \begin{tabular}{lcccccccc}
            \toprule
            \multicolumn{1}{c}{\multirow{2}{*}{\textbf{Model}}} & \multicolumn{1}{c}{\multirow{2}{*}{\textbf{LLM Size}}} & \multicolumn{4}{c}{\textbf{Temporal Grounding}} & \multicolumn{3}{c}{\textbf{Dense Captioning}} \\
            \cmidrule(lr){3-6} \cmidrule(lr){7-9}
             &  & R@0.3$\uparrow$ & R@0.5$\uparrow$ & R@0.7$\uparrow$ & mIoU$\uparrow$ & SODA\underline{~}c$\uparrow$ & CIDEr$\uparrow$ & METEOR$\uparrow$ \\
            \midrule
            VideoChat\cite{journals/corr/abs-2305-06355} & 7B & 8.8 & 3.7 & 1.5 & 7.2 & 0.9 & 2.2 & 0.9 \\
            Video-LLaMA\cite{conf/emnlp/ZhangLB23} & 7B & 6.9 & 2.1 & 0.8 & 6.5 & 1.9 & 5.8 & 1.9 \\
            Video-ChatGPT\cite{maaz2024acl} & 7B & 26.4 & 13.6 & 6.1 & 18.9 & 1.9 & 5.8 & 2.1 \\
            \rowcolor[rgb]{ .949,  .949,  .949} \textbf{PiTe (Ours)} & 7B & \textbf{30.4} & \textbf{17.8} & \textbf{7.8} & \textbf{22.0} & \textbf{5.1} & \textbf{21.7} & \textbf{5.8} \\
            \midrule
            \rowcolor[rgb]{ .949,  .949,  .949} \textbf{PiTe (Ours)} & 13B & \textbf{37.2} & \textbf{23.7} & \textbf{10.9} & \textbf{26.0} & \textbf{5.9} & \textbf{26.5} & \textbf{6.6} \\
            \bottomrule
        \end{tabular}
    }
\end{table}

\subsection{Main Result}

Table~\ref{tab:main_zsvqa} and~\ref{tab:main_an} present the comparative performance of the \textbf{PiTe} against state-of-the-art baselines on myriad video understanding datasets
.

\subsubsection{Question Answering.}

As illustrated in Table~\ref{tab:main_zsvqa}, \textbf{PiTe} consistently outperforms the state-of-the-art pure instruction-tuning baselines in terms of all metrics on all datasets.
Compared to the top-performing baselines in each dataset, \textbf{PiTe} exhibited notable improvements in the average question answering accuracy, achieving a maximum enhancement of $4.8$ and an average improvement of $3.7$.
For example, \textbf{PiTe} substantially improves accuracy by $64.9$ to $68.4$ compared to Video-ChatGPT\cite{maaz2024acl} in MSVD-QA dataset\cite{conf/mm/XuZX0Z0Z17}.
The results showcasing \textbf{PiTe}'s proficiency in video comprehension and its capacity to deliver contextually relevant responses according to the given instructions.

\subsubsection{Temporal Grounding.}

As depicted in Table~\ref{tab:main_an}, \textbf{PiTe} achieves state-of-the-art performance in the video temporal grounding task across all metrics as well, demonstrating improvements ranging from $18.9$ to $22.0$ in mIoU compared to Video-ChatGPT\cite{maaz2024acl}.
This clearly indicates that trajectory alignment greatly enhances the ability of capture events in temporal dimension for LVidLMs.
The incorporation of object trajectories in the temporal dimension of the trajectory matrix equips the model with a precise understanding of temporal event boundaries, thereby establishing a solid foundation for accurate event localization.

\subsubsection{Dense Captioning.}

The outcomes of the dense captioning task, as delineated in Table~\ref{tab:main_an}, reveal that \textbf{PiTe} consistent boost compared to all state-of-the-art baselines.
Particularly noteworthy is the substantial $15.9$ increase in the CIDEr metric\cite{conf/cvpr/VedantamZP15} when compared to Video-ChatGPT\cite{maaz2024acl}.
This underscores the significance of fine-grained alignment in both spatial and temporal dimensions through trajectories, implying that \textbf{PiTe} acquires more generalized and detailed representations to offer more sophisticated event descriptions and accurate event temporal boundaries.

\begin{table}[!t]
    \caption{Ablation study of the three-stage training strategy.}
    \label{tab:ablation}
    \centering
    \resizebox{\linewidth}{!}{
        \begin{tabular}{lccccccccc}
            \toprule
            \multicolumn{1}{c}{\multirow{3}{*}{\textbf{Method}}} & \multicolumn{2}{c}{\multirow{2}{*}{\textbf{MSVD-QA}}} & \multicolumn{7}{c}{\textbf{ActivityNet}} \\
            \cmidrule(lr){4-10}
             & \multicolumn{2}{c}{} & \multicolumn{4}{c}{Temporal Grounding} & \multicolumn{3}{c}{Dense Captioning} \\
            \cmidrule(lr){2-3} \cmidrule(lr){4-7} \cmidrule(lr){8-10}
             & Accuracy & Score & R@0.3 & R@0.5 & R@0.7 & mIoU & SODA\underline{~}c & CIDEr & METEOR \\
            \midrule
            \rowcolor[rgb]{ .949,  .949,  .949} \textbf{PiTe (Ours)} & \textbf{68.4} & \textbf{3.9} & \textbf{30.4} & \textbf{17.8} & \textbf{7.8} & \textbf{22.0} & \textbf{5.1} & \textbf{21.7} & \textbf{5.8} \\
            \quad w/o initialize & 68.2 & \textbf{3.9} & 22.8 & 10.5 & 4.6 & 17.1 & \textbf{5.1} & \textbf{21.7} & \textbf{5.8} \\
            \quad w/o trajectory & 68.1 & \textbf{3.9} & 23.9 & 12.8 & 5.7 & 17.4 & 5.0 & 21.4 & \textbf{5.8} \\
            \bottomrule
        \end{tabular}
    }
\end{table}

\subsection{Analysis}
\label{sec:ablation}

\subsubsection{Ablation Study.}

As reported in Table~\ref{tab:ablation}, we conduct ablation experiments on MVSD-QA\cite{conf/mm/XuZX0Z0Z17} for question answering and ActivityNet Captions\cite{conf/iccv/KrishnaHRFN17} for temporal grounding to verify the individual effects of the proposed contributions under the following settings:
(1) w/o initialize: we remove the initialization strategy that use weight of localization projector to initialize trajectory projector;
(2) w/o trajectory: we abandon the fine-grained alignment strategy via trajectory.

From the experimental results in Table~\ref{tab:ablation}, it can be observed that:
(1) Eliminating the initialization strategy for the trajectory projector in \textbf{PiTe} reduces the model's reasoning capabilities and temporal boundary awareness.
However, the performance in dense captioning generation remains consistent.
This observation suggests that the model maintains its basic ability in comprehending visual content under trajectory-guided training.
(2) The removal of the trajectory-guided training diminishes almost all the capability of \textbf{PiTe}, including dense captioning.
(3) Without trajectory-guided training, \textbf{PiTe} demonstrates superior performance compared to trajectory-guided training without the initialization strategy for the trajectory projector in temporal grounding.
This outcome highlights the difficulty of trajectory-guided training without initialization from a pre-trained localization projector, as the instability of parameters can impede the model's perception to accurately perceive visual temporal information.

\subsubsection{Exhibition.}

To better illustrate the video dialogue performance of \textbf{PiTe}, we present a qualitative example, as shown in Fig.~\ref{fig:example}.
The illustration from the upper portion of the figure demonstrates \textbf{PiTe}'s capability not only to provide precise responses to instruction queries but also to enhance the output with more detailed and accurate video information.
The example in the lower segment of the figure highlights the model's proficiency in understanding instruction and capturing event, enabling precise delineation of temporal boundaries within the video, despite the constraint of a 100-frame sampling limit.

\subsubsection{Impact of Tracking Point Quantity.}
\label{sec:n_point}

\begin{figure}[!t]
    \centering
    \subfloat[Examples of PiTe's video understanding capabilities.]{
        \label{fig:example}
        \includegraphics[width=0.43\linewidth,trim=0 0 0 0,clip]{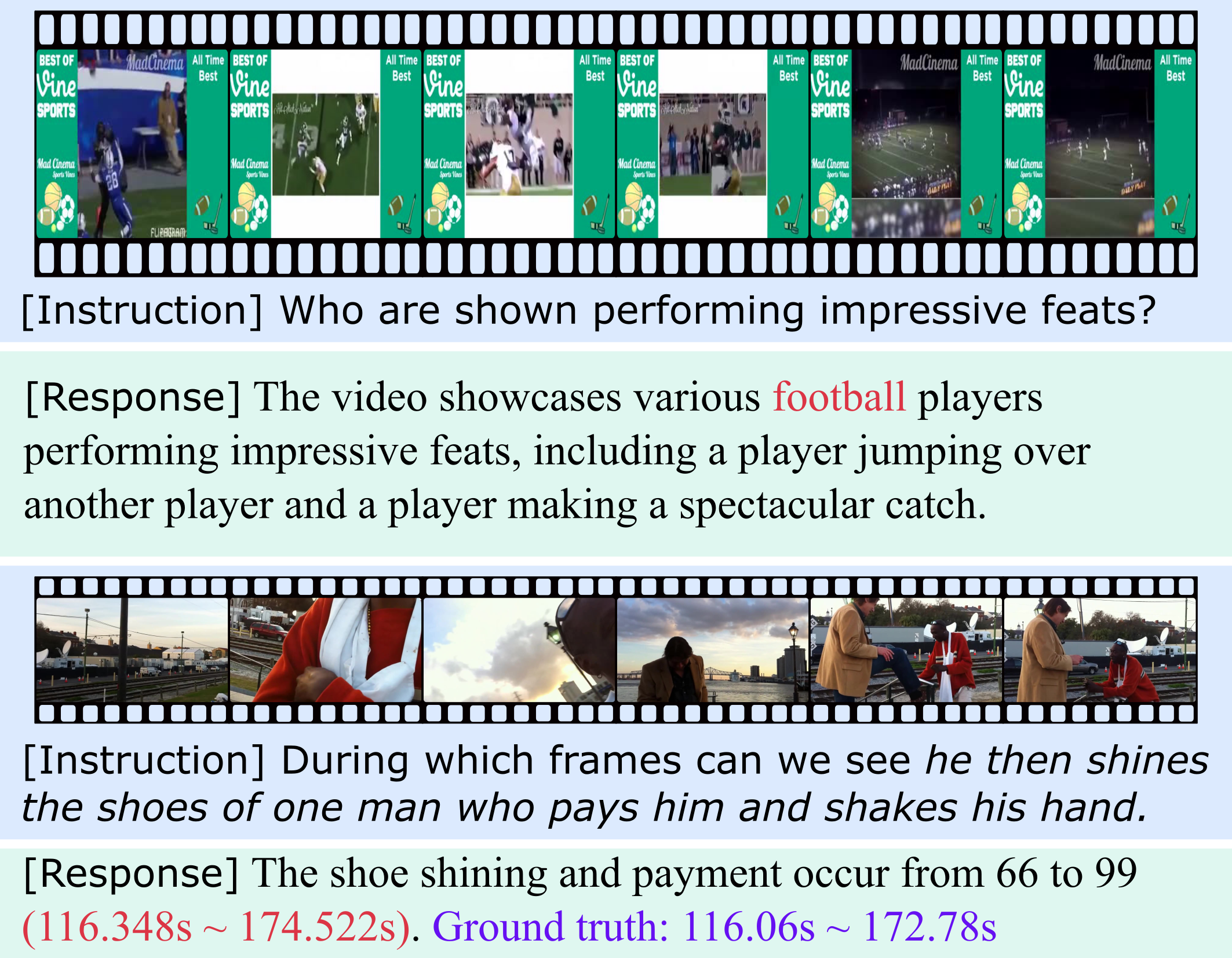}
    }
    \hfill
    \subfloat[Performance comparison for different tracking point quantity.]{
        \label{fig:n_point}
        \includegraphics[width=0.50\linewidth,trim=20 0 5 30,clip]{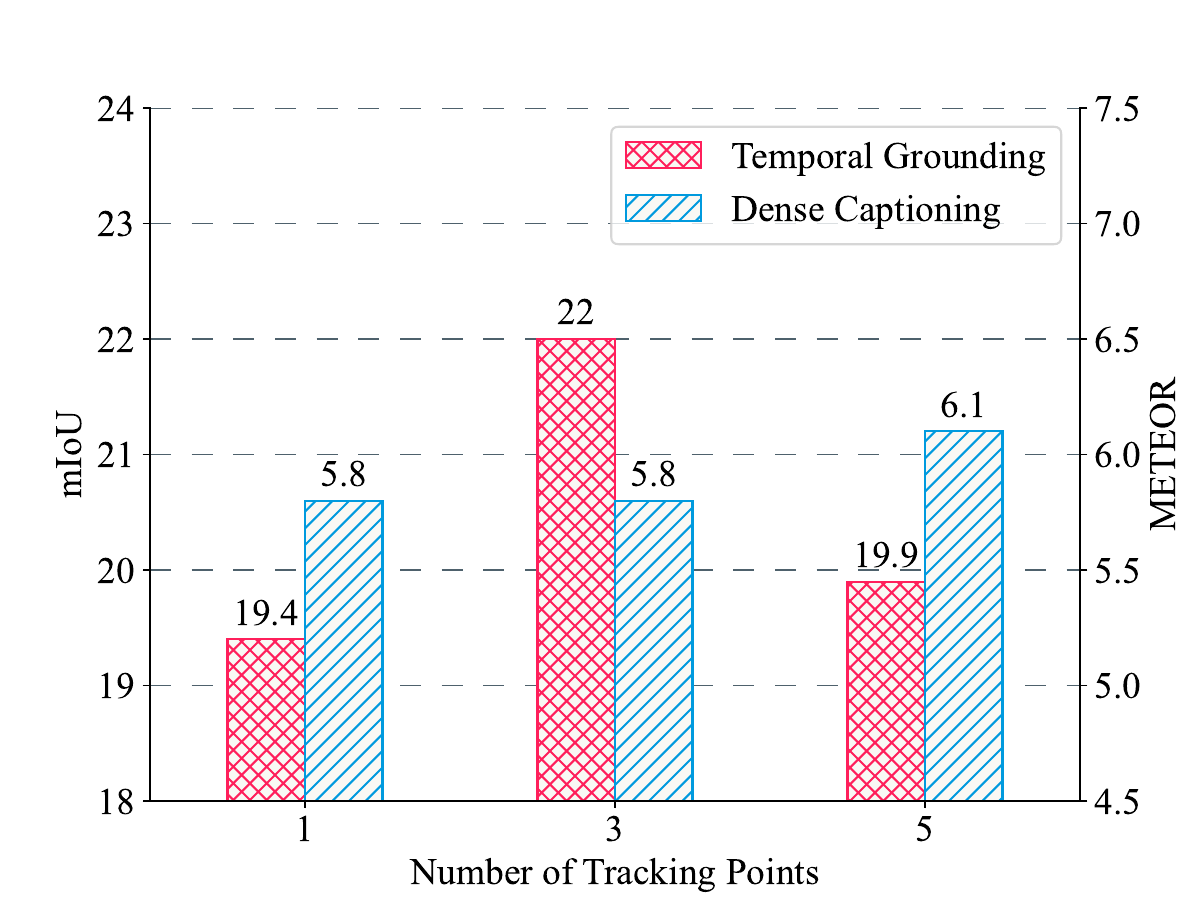}
    }
    \caption{PiTe's video understanding capabilities and performance comparison across varying tracking point quantities.}
    \label{fig:example_point}
    \vspace{-1.5ex}
\end{figure}

In Fig.~\ref{fig:n_point}, we vary the tracking point quantity $P$ in set of $\{1, 3, 5\}$.
The efficacy of dense captioning tasks tends to improve with an increase in tracking points.
However, it is observed that the temporal grounding task undergoes an initial substantial improvement, only to be succeeded by a rapid decline.
Less number of tracking points fails to accurately capture the object's geometry, thereby hindering the pixel-level cross-modal alignment guidance for the model.
Conversely, a higher quantity of points can enhance the model's comprehension of pure visual information; however, it also introduces noise to make training more challenging.
Overall, that the optimal value of $P$ may be different for different tasks, we set $P=3$ due to its performance maintain stability over multiple tasks.

\section{Conclusion}

In this paper, we focus on enhancing the performance of Large Video-Language Models (LVidLMs) by incorporating trajectory-based alignment across different modalities.
To achieve fine-grained alignment between video and language across spatial and temporal dimensions, we initially curate a comprehensive multi-modal object tracking dataset, PiTe-143k, using a fully automated annotation pipeline.
This dataset was developed to address the lack of large-scale video-language datasets that include multi-object moving trajectories.
Subsequently, we introduce a novel \textbf{Pi}xel-\textbf{Te}mporal (\textbf{PiTe}) alignment strategy that leverages trajectory-guided pre-training to address the inherent challenges faced by LVidLMs.
Through comparative analyses, we evaluate \textbf{PiTe} against state-of-the-art models and competitive baselines across various tasks in a zero-shot setting, including question-answering, temporal grounding, and dense captioning, showcasing the superior performance of \textbf{PiTe} with more sophisticated event descriptions and accurate event temporal boundaries.


%
%
\bibliographystyle{splncs04}
\bibliography{reference}
\end{document}